\documentclass[11pt]{article}
\usepackage[]{acl}

\usepackage{graphicx} 
\usepackage{tabularx}
\usepackage{listings}
\usepackage{float}
\usepackage{geometry}
\usepackage{amsmath}
\usepackage{algorithmicx}
\usepackage{appendix}
\usepackage{caption}
\usepackage{times}
\usepackage{soul}
\usepackage{booktabs}
\usepackage{multirow}
\usepackage{makecell}

\usepackage{comment}
\usepackage{todonotes}

\title{Know When To Stop:\\ A Study of Semantic Drift in Text Generation}

\author{Ava Spataru$^{1}$ \quad Eric Hambro$^{2\dagger}$ \quad Elena Voita$^{1}$ \quad Nicola Cancedda$^{1}$ \bigskip\\
  $^1$FAIR, Meta \quad 
  $^2$Anthropic \\
  {\tt \{avaspataru, lenavoita, ncan\}@meta.com}  \\ {\tt eric.hambro@gmail.com} 
	}

\date{}

\usepackage{xcolor}
\definecolor{darkgreen}{rgb}{0,0.5,0}
\definecolor{greyedout}{RGB}{96,96,96}
\newcommand{\redtext}[1]{\textcolor{red}{#1}}

\newcommand{\greentext}[1]{\textcolor{darkgreen}{#1}}

\definecolor{oracle}{HTML}{a61a12}
\definecolor{internal}{HTML}{2c8c11}

\newcommand\blfootnote[1]{%
  \begingroup
  \renewcommand\thefootnote{}\footnote{#1}%
  \addtocounter{footnote}{-1}%
  \endgroup
}

\begin{document}

\maketitle

\blfootnote{\textdagger Work done while at FAIR, Meta.}
\begin{abstract}

In this work, we explicitly show that modern LLMs tend to generate correct facts first, then ``drift away'' and generate incorrect facts later: this was occasionally observed but never properly measured. We develop a semantic drift score that measures the degree of separation between correct and incorrect facts in generated texts and confirm our hypothesis when generating Wikipedia-style biographies. This correct-then-incorrect generation pattern suggests that factual accuracy can be improved by \textit{knowing when to stop} generation. Therefore, we explore the trade-off between information quantity and factual accuracy for several early stopping methods and manage to improve factuality by a large margin. We further show that reranking with semantic similarity can further improve these results, both compared to the baseline and when combined with early stopping. Finally, we try calling external API to bring the model back to the right generation path, but do not get positive results. Overall, our methods generalize and can be applied to any long-form text generation to produce more reliable information, by balancing trade-offs between factual accuracy, information quantity and computational cost.

\end{abstract}

\section{Introduction}

Differently from the earlier approaches to generating natural language with explicit content planning~\cite{Mann1983AnOO,Reiter1997BuildingAN}, modern autoregressive language models make predictions token-by-token, without pre-established text structure. One of the consequences of this methodological shift is that newer models lack the capability of maintaining high-level structure throughout generation and overly focus on local coherence. This was noted in the form of repetition~\cite{fu2021theoretical} and semantic drift \cite{li-etal-2021-addressing-semantic}.

The term ``semantic drift'' emerged to describe the decrease in text generation quality when increasing generation length and has been classified as a sub-type of hallucinations~\cite{Ji_2023}. Before that, semantic drift (or topic shift) was briefly mentioned when talking about question generation~\cite{Zhang2019AddressingSD} and story generation \cite{wang2021sentence,sun2020improving}. In factual evaluation, recent works also mention a decline in factual accuracy for longer generations~\cite{min2023factscore,qian2023optimizing}. 
While quality decrease for longer generations hints at specific order in generation quality (high-quality first, low-quality later), this ordered behavior has not been neither formally defined nor thoroughly studied and measured.  In this work, we refer to ``semantic drift'' as the \textit{strength of the order} in generation quality and, for the first time, provide tools for understanding this phenomenon.

We propose to measure semantic drift by considering the change in truthfulness of a sequence of facts when a model generates a fact-rich text around a topic. Intuitively, we measure the \textit{degree of separation} between correct and incorrect facts in a paragraph:
if the model starts by generating correct facts and then switches to systematically generating incorrect ones, we consider this as a semantic drift. To quantify the severity of semantic drift, we use the FActScore task which provides correct/incorrect labels for individual facts~ \cite{min2023factscore}. We find that, indeed, several LLaMa2 variants have high semantic drift score: they tend to generate correct facts first, then ``drift away'' from the topic and generate incorrect facts later.

This correct-then-incorrect separation suggests that factual accuracy can be improved by stopping generation early. We show that even a simple method that encourages generating \texttt{EOS} leads to large improvements in factuality. 
We then propose to use resample-then-rerank pipelines where for each sentence, we generate several versions and choose the best based on sentence similarity measures. 
Compared to the baseline, this improves factual accuracy by almost $10\%$ (without shortening texts as with early stopping). This can also be combined with early stopping and allows for different informativeness-vs-factuality trade-offs. 
Finally, we ask: If the model drifts away during generation, could it be brought back onto a correct path by calling an external API? Sadly, this does not give noticeable improvements (at least, when working in the previously established settings).

Overall, we:
\begin{itemize}
    \item formally show that current LLMs tend to generate facts in a correct-then-incorrect manner;
    \item based on that, develop methods to improve factual correctness: simple early stopping and more complex resample-then-rerank;
    \item find that API calls help little to none.
\end{itemize}

Our methods offer a practical compromise, balancing computation with performance, and build a foundation for further research. Importantly, they are directly applicable to any probabilistic auto-regressive language models.

\section{Definition of Semantic Drift }
Since the term ``semantic'' drift has been used with various meanings,  we felt the need for a unifying definition, which we state below.
\begin{algorithmic}

\State \textit{Semantic drift describes the phenomenon wherein generated text diverges from the subject matter designated by the prompt, resulting in a \textbf{growing} deterioration in relevance, coherence, or truthfulness.}
\end{algorithmic}

\noindent Semantic drift results in a loss of three textual characteristics (see examples in Appendix~\ref{app:sd}):
\begin{enumerate}
    \item Loss in \textbf{coherence}, which leads to issues with clarity, logical flow, and self-consistency;
    \item Loss in \textbf{relevance}, which refers to the inclusion of irrelevant or redundant content;
    \item Loss in \textbf{truthfulness}, which refers to the inclusion of hallucinated content or content inconsistent with world knowledge. 
\end{enumerate}

\subsection{Semantic Drift Score} \label{sec:driftdef}
\par To quantify the severity of semantic drift, we define a new scoring method, semantic drift (SD) score.
To calculate this score for a paragraph $P$, we take individual atomic facts along with their labels (1 for supported facts and 0 otherwise). Let~$N$ be the total number of facts, $s_i$ be the label for the fact $i \in [0,N)$, and $m$ be a  hyperparameter. Then we define the SD score as: 
\begin{align*}
    &SD_m(P) = \max\limits_{k}  \frac{1}{2}\cdot SD_m(P,k) \\
    &SD_m(P,k) = 
        \begin{cases}
        \begin{aligned}
            &0, \hspace{0.4cm}\text{ if }(N\!-\!k\!<\!m) \text{ or } (k\!<\!m) \\
            & \hspace{1cm} \text{ \hspace{4.4em} or }   (N\!<\!2m), \\
            &\frac{ \sum_{i=0}^{k-1} s_i}{k} + \frac{\sum_{i=k}^{N-1}(1\!-\!s_i)}{N-k} \text{ else.}
        \end{aligned}
        \end{cases}
\end{align*}

\noindent
The position $k$ at which this maximum is reached represents the position with highest average between (1)~proportion of supported facts to the left of position $k$ and (2)~proportion of not-supported facts to the right of position $k$. We will refer to $k$ as the \textit{drift point}. Parameter $m$ controls the range of $k$, meaning that we only consider splits that have more than $m$ facts on either side of $k$.

\begin{figure}[t!]
  \includegraphics[width=\linewidth]{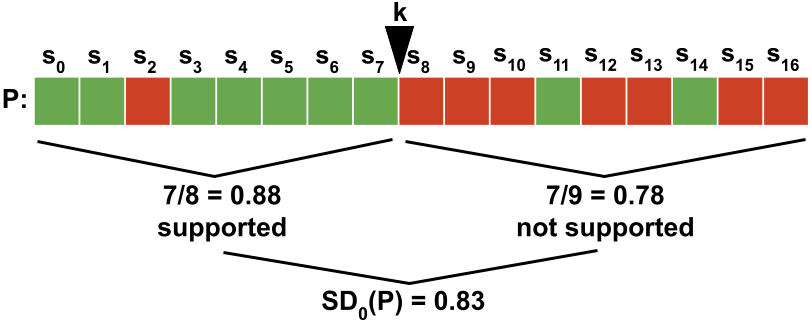} 
  \caption{A visual example of calculating semantic drift (SD) score for paragraph $P$. The position which best splits the paragraph is $k=8$. The proportion of supported facts to the left is 0.88 and the proportion of not-supported facts to the right is 0.78, giving an average of 0.83. The other positions all have lower SD scores, therefore the SD score of paragraph $P$ is 0.83.}
  \label{fig:sd_score_def}
\end{figure}

\par Intuitively, we measure the \textit{degree of a separation} between correct and incorrect facts in a paragraph: the SD score is high when a text is largely correct before the drift point and largely wrong after (Figure~\ref{fig:sd_score_def}). E.g., a paragraph with an SD score of 1 would have 
all correct facts first and all the incorrect facts later. For a paragraph in which facts are either wrong or correct without any clear separation, we would expect an SD score around 0.5.

\section{Identifying Semantic Drift}

\par In our experiments, we rely on the FActScore task \cite{min2023factscore}. This task identifies all aspects of semantic drift and scores individual facts from a text as either correct or incorrect. A fact is correct if it is supported by external knowledge and therefore truthful. Since two facts that contradict each other cannot be simultaneously correct, correct facts are also coherent. Moreover, facts are verified in context, meaning that a fact is correct only if it is relevant to the context. Appendix~\ref{app:sd} shows examples of scoring for semantic drift types.

\subsection{Setting} \label{sec:task}

\paragraph{Task.} The FActScore task focuses on Wi\-ki\-pe\-dia-style biographical passages: they are generally fact dense, and the individual facts can be reliably verified~\cite{wadden-etal-2020-fact,thorne-etal-2018-fever}. 
The task consists of 3 steps: (1)~generating biographical paragraphs for 500 entities, (2)~extracting ``atomic facts'' from the paragraphs, and (3)~scoring the truthfulness of paragraphs by verifying all atomic facts against a knowledge source. The FActScore itself is the precision of atomic facts aggregated over the 500 samples.

 \paragraph{Pipeline.} We let LLaMa2-70B generate a biographic paragraph, using the same prompt as \citet{Manakul2023SelfCheckGPTZB}: \textit{``This is a Wikipedia article about [entity]. [entity]''}.\footnote{Example prompt: ``\textit{This is a wikipedia article about Bob Marley. Bob Marley}''. The model has to continue generation.}  Each generated paragraph is then passed through the FActScore pipeline to identify and verify atomic facts. We modify the original FActScore pipeline to rely on LLaMa2-70B-Chat (rather than InstructGPT) and validate using human annotations (Appendix~\ref{sect_apdx:adapt_factscore_to_llama}).

\subsection{Semantic Drift in LLaMa2-70B}\label{sec:driftproof}

\begin{figure}
\centering
    \includegraphics[width=6cm]{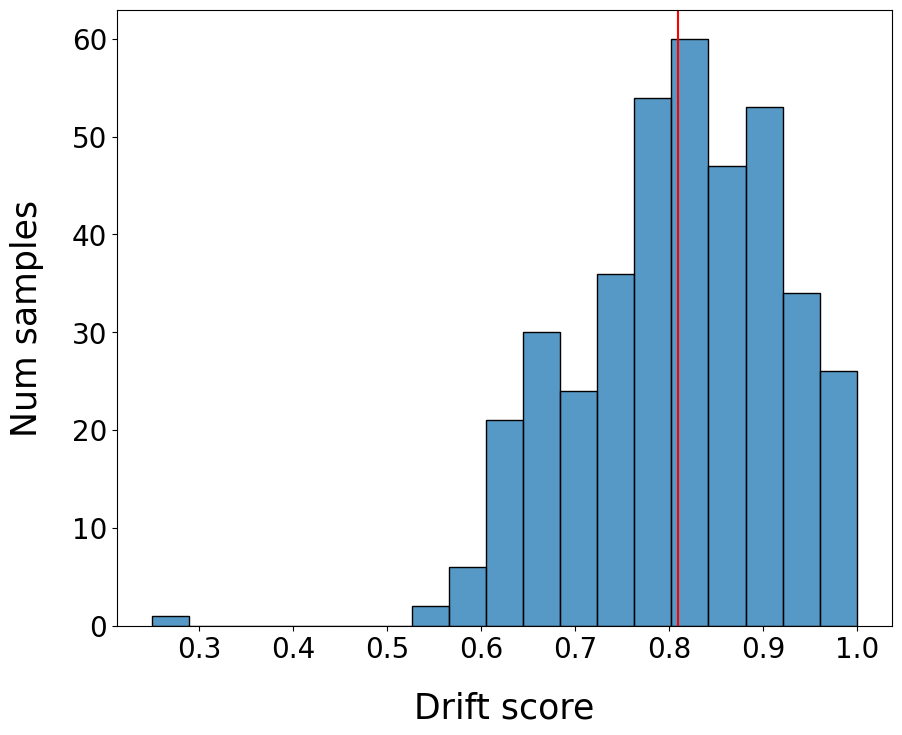}
    \vspace{-2ex}
    \caption{Distribution of Semantic Drift Score (after filtering) in paragraphs generated by LLaMa2-70B (sampling: temperature=0.6, top-p=0.9).}
    \label{fig:llama2-distrib}
    \vspace{-2ex}
\end{figure}

For paragraphs generated by LLaMa2-70B, we got an average SD score of \textbf{0.78} when considering all 500 examples and the score of \textbf{0.8} when filtering out completely correct and incorrect samples.\footnote{Filtered 20 completely incorrect and 21 completely correct samples, remained with 458 generated paragraphs with an average of 47 facts per paragraph.} Figure \ref{fig:llama2-distrib} shows the distribution of SD scores.

\paragraph{Semantic drift is high.} 

Semantic drift score of~$0.8$ is very high: it means that there is a \textit{significant separation between correct and incorrect facts in most paragraphs}, and thus model generations ``drift away'' at some point during generation. To ensure that the high semantic drift score is not just chance, we conduct a statistical significance test. We estimate the probability that a random permutation of facts would result in the same SD score or higher. We find that samples we identified as drifting have an average probability of \textless 0.02. For more details, please refer to Appendix \ref{app:a:stat}.

\paragraph{Drift starts early.}
When looking at the number of correct facts, we noticed that generations are largely wrong, and the drift starts early. For example,
only a small portion of paragraphs has at least 10 correct facts before the first wrong fact (34 paragraphs, $< 7\%$). For $37\%$ of all paragraphs, the drift point is in the first $10\%$ of facts. 

\paragraph{Our observations are reliable.} In Appendix \ref{app:A}, we show that decoding strategy and truncation parameter only slightly impact the SD score (hence, our observations). We conclude that LLaMa2-70B shows statistically significant high SD score in more than $40\%$ of generated paragraphs. 

\subsection{Semantic Drift in Other Models}
To strengthen our observations, we extend our experiments to other well-established LLMs.

\paragraph{Setting.} We consider LLaMa2-70B-Chat~\cite{touvron2023llama}, Falcon~\cite{falcon40b} and GPT~\cite{openai2023gpt4}.  These models are both text and chat completion models. For text completion, we use the same prompt as in Section~\ref{sec:task}. For chat completion, we use ``\textit{Tell me a bio of }''. 

\paragraph{Results.} From Table~\ref{tab:other_models} we see that semantic drift is high for all models. This confirms our hypothesis: models start with correct facts, then ``drift away''. While the GPT models perform considerably better on the FActScore* task, they still have high SD score and could therefore benefit from our error mitigation strategies from Section~\ref{sec:mitigate-all}.

\begin{table}[t]
  \small
    \begin{tabular}{l|c|c|c|c}
        \toprule
          &  \bf facts&  \bf No &  \bf FAct & \bf SD \\
          \bf Model &  \bf /gen&  \bf ans-&  \bf Score*& \bf Score\\
           &   &  \bf wer &  \bf (\%)& \bf (\%)\\
         \toprule
         Llama2-70B-chat&  50.55&  1&  41.72& 77.06\\
         \hline
         Falcon-7B& 41.84 & 6 & 24.64 & 76.81 \\
         Falcon-40B& 49.23 & 4 & 25.88 & 77.38 \\
         \hline
         text-davinci-003&58.27 & 2 & 38.09 & 77.21 \\
         GPT 3.5& 67.82 & 1 & 45.96 & 79.49 \\
         GPT 4& 48.31 & 1 &53.54 &78.12 \\
         \bottomrule
    \end{tabular}
    \caption{Results for various models when generating 500 biographical paragraphs.  ``No answer'' is the number of paragraphs when the model produced no facts.}
    \label{tab:other_models}
    \vspace{-2ex}
\end{table}

\section{Analysis}
Let us now understand the potential causes of semantic drift and analyze LLaMa2-70B generations  both quantitatively  and qualitatively.

\subsection{Quantitative Analysis}
We analyze the distribution of semantic drift by multiple factors:  (i)~person popularity, (ii)~paragraph length and drift position, (iii)~model scale.

\begin{figure}
    \centering
    \includegraphics[scale=0.25]{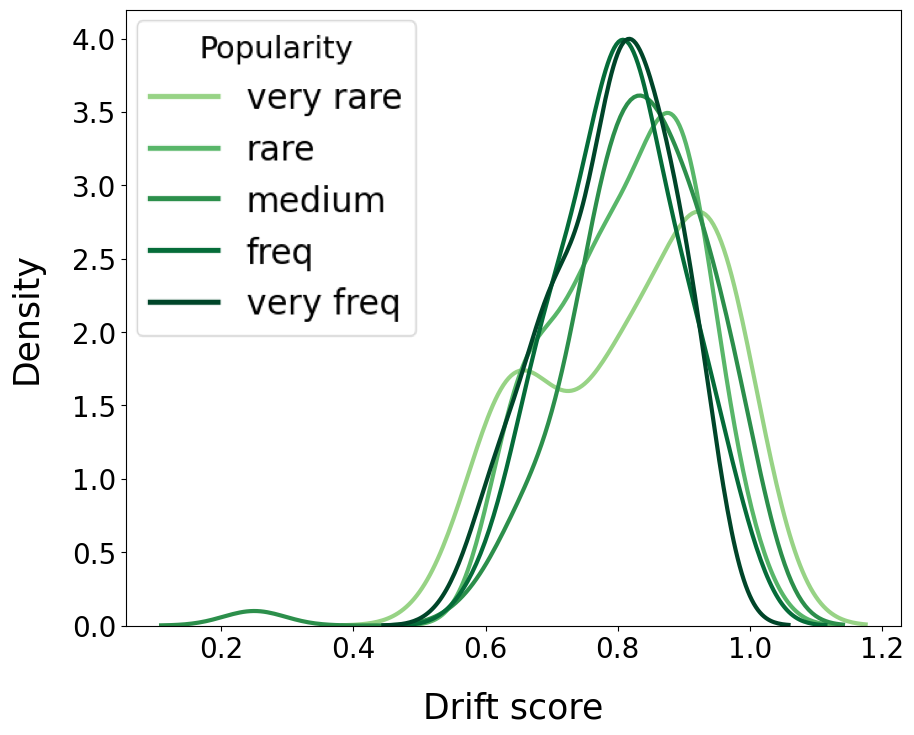}
    \caption{Semantic drift score density plot for person popularity classes. LLaMa2-70B.}
    \label{fig:freq}
\end{figure}

\paragraph{Person popularity.} 
We hypothesize that semantic drift score might be affected by the popularity of a bio's object in a typical dataset.
Figure~\ref{fig:freq} shows the distribution of SD scores by prevalence class, from ``very rare'' to ``very frequent''.\footnote{The prevalence labels come from the FActScore dataset.} We see that for very frequent entities, the semantic drift score is distributed normally. As the entities become less frequent, the distribution starts turning into a bimodal distribution. This could be because for rare entities, the model either generates a few facts well and then drifts away (resulting in high drift score), or has a generally murky knowledge about the entity and generates both wrong and correct facts together (resulting in low drift score).

\paragraph{Paragraph length and drift position.} We find no correlation between paragraph length, drift score and relative drift position. However, we do note that the distribution of relative drift position is distinctly U-shaped, with more paragraphs drifting in the first $10\%$ of generated facts than in the last $10\%$. We apply truncation as described in Section~\ref{sec:driftdef} and note that the distribution of drift position is still peaking in the first $10\%$ of generated facts. For more details, see Appendix \ref{app:a:prune}.

\paragraph{Model scale.} Table~\ref{tab:scaling_laws} shows results for the same pipeline with two smaller LLaMa2 models. We find that while increasing parameter size clearly improves factuality of generated text, all three model sizes show similar SD scores:  semantic drift is high  regardless of scale.

\begin{table}[t]
  \small
    \begin{tabular}{c|c|c|c|c}
        \toprule
         \bf Num &  \bf facts&  \bf No &  \bf FAct Score* & \bf SD Score\\
          \bf params&  \bf /gen&  \bf answer&  \bf (\%)& \bf (\%)\\
         \toprule
         7B&  34.80&  4&  34.77& 76.29\\
         13B&  33.04&  1&  39.36& 76.83\\
         70B&  43.89&  1&  44.56& 78.07\\
         \bottomrule
    \end{tabular}
    \caption{FActScore* and SD Score for different model sizes of LLama2 when generating 500 biographical paragraphs.  ``No answer'' shows the number of paragraphs for which the model produces no facts.}
    \label{tab:scaling_laws}
\end{table}

\subsection{Qualitative Analysis}\label{sec:qualanalysis}

\begin{table*}[ht]  
  \small
  \begin{tabularx}{\textwidth}[b]{  >{\hsize=0.14\hsize}X|X  }
    \toprule
    \textbf{Drift Type} & \textbf{Examples}\\ \toprule
      Repetition & Suthida is the \greentext{Queen of Thailand}. She is the \greentext{wife of King Vajiralongkorn}. She was \greentext{born on 3 June 1978}. She is the \redtext{first} \greentext{wife of King Vajiralongkorn}. She was \greentext{born in Hat Yai, Songkhla, Thailand}. She is the \redtext{daughter of Saman Bajrasudha and his wife, Nui}. She has two brothers and one sister. She is a \greentext{former flight attendant}. She is a \redtext{former actress}. She is a \redtext{former model}. She is a \redtext{former beauty queen}. She is a \redtext{former Miss Thailand}. She is a \redtext{former Miss Universe}. She is a \redtext{former Miss World}. \\ \hline

        Making up information & Steve Cummins is a \greentext{professional rugby league footballer} of the \redtext{1990s and 2000s}. A \redtext{Great Britain international representative centre}, he has played in the Super League for the \redtext{Wigan Warriors} and the \redtext{Warrington Wolves}, and in the NRL for the \redtext{Brisbane Broncos} and the \redtext{Cronulla-Sutherland Sharks}. \\ \hline

      Plausible information & Matthew Perry is an \redtext{American} \greentext{actor and comedian}. He is \greentext{best known for his role as Chandler Bing} on the \greentext{NBC television sitcom Friends},  \greentext{which ran from 1994 to 2004}.
Matthew Perry was  \greentext{born in Williamstown, Massachusetts},  \greentext{on August 19, 1969}. His  \greentext{mother, Suzanne Marie Morrison}, is a  \greentext{Canadian journalist and former press secretary to Canadian Prime Minister Pierre Trudeau}. His  \greentext{father, John Bennett Perry}, is an  \greentext{American actor and former model}. Perry has  \greentext{two older sisters}, Caitlin and Emily.
Perry was  \greentext{raised in Ottawa, Ontario}, and  \greentext{attended Rockcliffe Park Public School} and \greentext{Ashbury College}. He then \redtext{studied at the University of Southern California}, where he was a \redtext{member of the Sigma Chi fraternity}.
  \\ \bottomrule

  \end{tabularx}
  \caption{Examples of types of semantic drift described in section \ref{sec:qualanalysis}. The ``Suthida'' example gets stuck on a loop of false facts after the phrase ``former flight attendant''.  The ``Steve Cummins'' example shows one correct fact followed by many made-up ones. We classified the ``Matthew Perry'' example as plausible information, since Perry intended to enroll at the University of
Southern California; same phrasing for fraternity appears on Perry's father's Wikipedia page; and there was a Phil Perry attending Sigma Chi fraternity. }
  \label{tab:example-table}
  \vspace{-2ex}
\end{table*}

We looked at examples of biographies generated by LLaMa2-70B with high SD score and identified at least three potential categories of semantic drift (Table~\ref{tab:example-table}):
\begin{enumerate}
    \item \textbf{Repetition.} The clearest and easiest type of semantic drift happens when the model ``gets stuck'' in a loop of identical or similar facts. 
    \item \textbf{Making up information}. This type of drift happens when the model can generate a few correct generic facts (such as occupation), but makes up all subsequent information. 
    \item \textbf{Plausible information.} This is perhaps the most interesting and subtle type of drift. It happens when the model generates a good lengthy biographic paragraph, but towards the end begins adding information which is indirectly relevant and incorrectly attributes that information to the subject. 
\end{enumerate}

\section{Factual Accuracy and Uncertainty}

Knowing that the model is largely correct up to some point (to be precise, the semantic drift point) and largely wrong after gives us an opportunity to improve generation quality. Specifically, if we can detect semantic drift during inference, we can stop generation (hence, improve its quality)~-- we will do this in Section~\ref{sec:mitigate-all}. But before that, let us check whether there are metrics that, during generation, can indicate that the model is drifting away.

Previous work on alle\-vi\-ating hallucinations for various NLP tasks, such as machine translation, abstractive summarization and long-form question answering, showed that hallucinations are well-calibrated with model uncertainty~\cite{lin2022teaching,kadavath2022language,liu2022tokenlevel,guerreiro-etal-2023-looking,Manakul2023SelfCheckGPTZB}. Here, we check whether uncertainty metrics correlate well with factual accuracy; for this, we use all sentences in the generated paragraphs (4516 sentences in total). 

\subsection{Considered Uncertainty Metrics}
\label{sect:uncertainty_metrics}

\paragraph{Intrinsic metrics:} We consider entropy of token probability distributions (averaged within a sentence), variance in entropy of token probability distributions (averaged within a sentence), negative log likelihood of the sentence. These metrics were used before in \citet{lin2022teaching,Manakul2023SelfCheckGPTZB, liu2022tokenlevel}. 

\paragraph{Intrinsic, averaged over samples:} to reduce noise in the metrics above, we sample each sentence 5 times and average the intrinsic uncertainty metrics over these samples.

\paragraph{Sampling-based (sentence similarity) metrics:} SelfCheck-BERT\-Score, SelfCheck-MQAG and SelfCheck-ngram (1, 5 and 10) from \cite{Manakul2023SelfCheckGPTZB}.  SelfCheck-BERTScore\footnote{From here onwards, ``SC-BERTScore'' for short.} assigns a unique score to a sentence, signifying how factual that sentence is (0 = factual, 1 = non-factual). To calculate the score, we sample N new paragraphs for each biography: $P_{1}$, $P_{2}$ and $P_{N}$. For each sentence $S$, we get the most similar sentence $S_{i}$ in each paragraph $P_{i}$, by considering maximum BERTScore. The SC-BERTScore is then calculated as $1-$ avg[BERTScore$(S, S_i)$]. For more details, see the original paper \cite{Manakul2023SelfCheckGPTZB}.

\subsection{Results}

\par We find that intrinsic uncertainty metrics have little correlation with factual accuracy and that averaging across samples does not improve this. Differently, sampling-based uncertainty metrics give much higher correlation scores; highest score gives SC-BERTScore with a Pearson correlation coefficient of -0.41 (Appendix~\ref{app:B}).

\section{Mitigating Factual Errors} 
\label{sec:mitigate-all}

As we explained above, the presence of semantic drift suggests that 
factual accuracy can be improved within the same generated paragraphs simply by shortening them. Therefore, we first consider several criteria to stop generation early. Next, we try resample-then-rerank pipeline, as well as calling API tools. We compare these methods through the lens of factuality vs informativeness trade-off (Figure~\ref{fig:compare_ending}).

\begin{table*}[ht]
\centering
  \small
  \begin{tabular}{lc|c|c|c|c|c|c|c}
    \toprule
    \multirow{2}{*}{\textbf{Method}} & \multirow{2}{*}{\textbf{Stop at}} & \multirow{2}{*}{\textbf{facts/gen}} & \multicolumn{1}{c|}{\multirowcell{2}{\textbf{No} \\  \textbf{ans.}}} & \multicolumn{1}{c|}{\multirowcell{2}{\textbf{FActScore*} \\  \textbf{(\%)}}} & \multicolumn{1}{c|}{\multirowcell{2}{\textbf{Recall} \\  \textbf{(\%)}}}& \multicolumn{1}{c|}{\multirowcell{2}{\textbf{SD Score} \\  \textbf{(\%)}}} & \multicolumn{2}{c}{\textbf{Flops*}} \\

    \textbf{} & \textbf{} & \textbf{} & \textbf{} & \textbf{} & \textbf{} & \textbf{} & \textbf{int.} & \textbf{ext.}  \\
    \toprule
    \bf Baseline & max tokens & 43.89 & 1 & 44.56 & - & 78.07 & $1e16$ & 0\\
    \hline
    \multicolumn{7}{l}{\bf Early stopping}\\
    \ \ \ \ \textcolor{oracle}{oracle} & \textcolor{greyedout}{drift point} & \textcolor{greyedout}{10} & \textcolor{greyedout}{1} & \textcolor{greyedout}{81.68}& \textcolor{greyedout}{41.76 } & \textcolor{greyedout}{47.94} & \textcolor{greyedout}{$2e15$} & \textcolor{greyedout}{0}  \\
    \cmidrule{2-8}
     \ \ \ \ EOS & \texttt{EOS} in top 5 & \colorbox{oracle!20}{14.47} & \colorbox{oracle!5}{3} & \colorbox{internal!15}{57.96} & \colorbox{oracle!20}{42.71} & \colorbox{internal!10}{74.20} & \colorbox{internal!50}{$5e15$} & 0  \\
     
     & \texttt{EOS} in top 10 & \colorbox{oracle!50}{\phantom{0}5.39} & \colorbox{oracle!20}{13} & \colorbox{internal!40}{70.29} & \colorbox{oracle!40}{18.90} & \colorbox{internal!35}{63.81} & \colorbox{internal!40}{$1e15$} & 0  \\
    \cmidrule{2-8}
     \ \ \ \ SC-BERT & SC-BERT incr. \textgreater 0.7 & \colorbox{oracle!15}{17.65} &1 & \colorbox{internal!20}{64.76} & \colorbox{oracle!10}{58.44} & \colorbox{internal!25}{66.87} & \colorbox{oracle!15}{$6e16$} & \colorbox{oracle!5}{$3e16$} \\
     
     & SC-BERT  incr. \textgreater 0.5 & \colorbox{oracle!25}{13.39} &1 & \colorbox{internal!30}{67.63}& \colorbox{oracle!20}{46.30}& \colorbox{internal!30}{65.20} & \colorbox{oracle!13}{$6e16$} & \colorbox{oracle!5}{$2e16$} \\
     
     & SC-BERT  incr. \textgreater 0.3 & \colorbox{oracle!30}{\phantom{0}9.66} & 1 & \colorbox{internal!40}{70.24}& \colorbox{oracle!30}{34.69}& \colorbox{internal!40}{\textbf{61.90}} & \colorbox{oracle!10}{$5e16$} & $1e16$ \\
    \hline
    \multicolumn{7}{l}{\bf Reranking (SC-BERT)}\\
     & max tokens  & 40.21 &1 & \colorbox{internal!10}{53.27} & - & \colorbox{internal!10}{74.84} & \colorbox{oracle!30}{$1e17$} & \colorbox{oracle!50}{$3e17$}\\
     
     & SC-BERT  incr. \textgreater 0.7 & \colorbox{oracle!5}{22.75} &1 & \colorbox{internal!20}{63.72}& \colorbox{oracle!5}{\textbf{67.67}} & \colorbox{internal!20}{69.15} & \colorbox{oracle!30}{$1e17$} & \colorbox{oracle!30}{$1e17$} \\
     
     & SC-BERT  incr. \textgreater0.5 & \colorbox{oracle!15}{17.12} & 1 & \colorbox{internal!30}{67.18} & \colorbox{oracle!10}{53.69} & \colorbox{internal!25}{67.49} & \colorbox{oracle!15}{$9e16$} & \colorbox{oracle!30}{$1e17$}  \\
     
     & SC-BERT  incr. \textgreater 0.3 & \colorbox{oracle!30}{11.64} & 1 & \colorbox{internal!40}{\textbf{71.11}} & \colorbox{oracle!20}{38.64} & \colorbox{internal!35}{63.94} & \colorbox{oracle!15}{$5e16$} & \colorbox{oracle!15}{$6e16$}  \\
    \hline

    \multicolumn{7}{l}{\bf API call}\\
    \ \ \ \ one QA call & max tokens & 42.26 & 1 & \colorbox{oracle!10}{43.93} & - & \colorbox{oracle!10}{80.02} & \colorbox{oracle!5}{$3e16$} & $1e10$ \\
    
    \ \ \ \ $\infty$ QA calls & max tokens & \colorbox{oracle!10}{19.36}& \colorbox{oracle!50}{54} & \colorbox{internal!10}{54.42} & - & \colorbox{internal!2}{77.43} & $1e16$ & $3e10$ \\
    \bottomrule
  \end{tabular}
  \caption{FActScore* and SD score for LLaMa2 70B with generation strategies and early stopping methods from Section \ref{sec:mitigate-all}, based on \textit{eos\_top\_k}, SelfCheck-BERTScore($N=3$), question answering calls or the oracle @drift point. Recall shows \%correct facts left from baseline. ``No ans'' shows number of paragraphs (out of 500) with no facts. ``Flops*'' approximates the total number of (internal and external) floating point operations.  }
  \label{tab:eosbert}
  \vspace{-2ex}
\end{table*}

 \subsection{Early Stopping}
 \label{sect:early_stopping}

 We consider several early stopping methods.

\paragraph{Oracle: at drift point.} This method stops generating at the drift point (Section~\ref{sec:driftdef}). While this cannot be achieved at inference time (finding the drift point requires ground truth that is not available at test time), this method gives us a theoretical upper bound of factual accuracy for early stopping methods and a reference point for other methods.

\paragraph{Incentivizing EOS.} As a naive baseline, we encourage the model to end the generation early by producing the \texttt{EOS} token whenever this token
is in the top-k predicted tokens.

\begin{figure}[t]
    \centering
    \includegraphics[width=7.5cm]{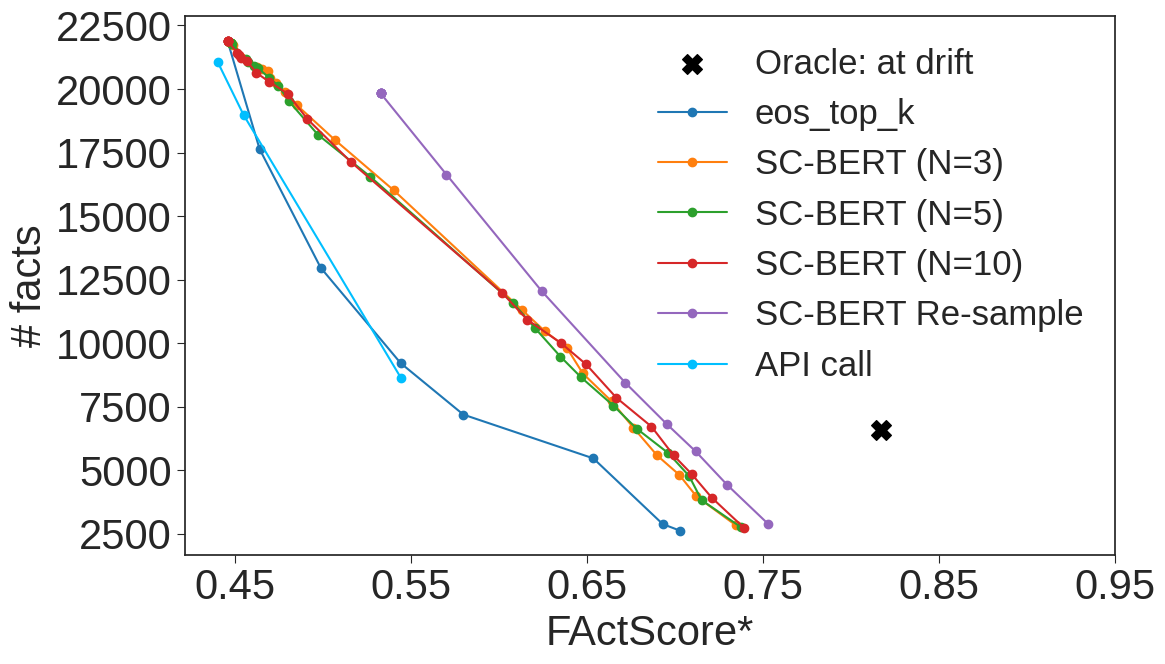}
    \vspace{-1ex}
    \caption{Trade-off between informativeness (y-axis) and factuality (x-axis) for proposed generation strategies; average over 500 biographical paragraphs.}
    \label{fig:compare_ending}
    \vspace{-2ex}
\end{figure}

\paragraph{Using sentence similarity.} Inspired by the correlation results between sentence similarity metrics and factual accuracy in Section~\ref{sect:uncertainty_metrics}, we also consider early stopping based on decline in consistency. For this, we: 
\begin{algorithmic}
  \State  \textbf{Step 1:} Compute SC-BERTScore for the original generated biographic paragraphs;
  \State \textbf{Step 2}: For each paragraph, calculate the percentage of increase in this score from sentence $S_{i}$ to $S_{0}$;
  \State \textbf{Step 3:} If this percentage is more than a threshold~ $T$ (i.e., consistency declined), stop the generation right before sentence $S_{i}$.
\end{algorithmic}
Here, $T$ controls how much information should be traded for factuality: a low~$T$ will result in shorter generations with higher factuality. Note also that SC-BERTScore depends on the number of paragraph samples $N$ which controls the accuracy of the scoring. 
 Since using $N>3$ does not give noticeable improvements (Figure~\ref{fig:compare_ending}), we use $N=3$.

\subsubsection{Results: Early Stopping Helps}

The results are shown in Table~\ref{tab:eosbert}. As expected, oracle (stopping at drift point) is the best: it has the highest factuality and the lowest drift score. Other early stopping methods also improve quality quite a lot and can achieve over $70\%$ accuracy (vs $44\%$ for the baseline) and low semantic drift score of $62$ (vs $78$ for the baseline). Naturally, stopping early leads to information loss and fewer generated facts overall. Therefore, we can compare different methods only in settings where, on average, they generate the same number of facts. Comparing \texttt{EOS}-in-top-5 with SC-BERTScore~($0.5$), we see that SC-BERTScore is better: for roughly the same number of generated facts per paragraph (13-14), it gives $10\%$ higher accuracy and lower SD score. In terms of flops, however, \texttt{EOS}-based stopping is an order of magnitude more efficient.\footnote{For SC-BERTScore stopping, the computation is two-fold: sampling four paragraphs instead of one and computing SC-BERTScore (three passes through RoBERTa-Large \cite{liu2019roberta}). This results in an order of magnitude more flops than the baseline.}

\subsection{Re-Sample, Then Rerank} 
\label{sec:rerank}

In addition to early stopping, SC-BERTScore can be used in resample-then-rerank pipelines
typical for alleviating hallucinations in machine translation~\cite{guerreiro-etal-2023-looking,dale-etal-2023-detecting}.

\paragraph{Method.} For each biography, we generate one sentence at a time. For each sentence, we generate 5 options (using same decoding strategy, only different seeds) and choose the one which (1)~has not appeared before in the paragraph; (2)~has minimum SC-BERTScore. We generate sentences until no options satisfy condition (1) or we have reached the maximum number of tokens. 

\paragraph{Results.} When stopping at the maximum number of tokens, this approach improves the baseline by $8.71\%$. This is expected: similar approaches improve e.g. machine translation quality by a large margin~\cite{guerreiro-etal-2023-looking,dale-etal-2023-detecting}. When combining reranking with the early-stopping, we get same factuality as the corresponding early stopping, but with more generated facts. For example, for the same factuality of around $67\%$ we generate $13.4$ facts with early stopping (SC-BERT, $T=0.5$) but $17.1$ facts when combining it with reranking. Sadly, this improvement comes with a large increase in flops.

\subsection{Calling Question Answering API}
\label{sect:api_call}

In this section, we ask: \textit{If the model drifts away during generation, could it be brought back onto a correct path by calling an external API?}

\paragraph{Method.}
To answer this question, we use 1-shot learning to allow LLaMa2-70B asking questions at inference time (as in Toolformer, \citet{schick2023toolformer}). The model makes calls to Atlas, which is a retrieval-augmented model with 11B parameters~(\citet{izacard2022atlas}, example inference in Appendix \ref{app:D}). We estimated the computation cost in Table \ref{tab:eosbert} by approximating the cost of an API call as the cost of one pass through the model. We find that when allowing for multiple calls, generations are shorter and therefore require fewer passes through LLaMa2. Therefore, overhead for adding QA calls is small.

\begin{figure}
    \centering
    \includegraphics[width=\linewidth]{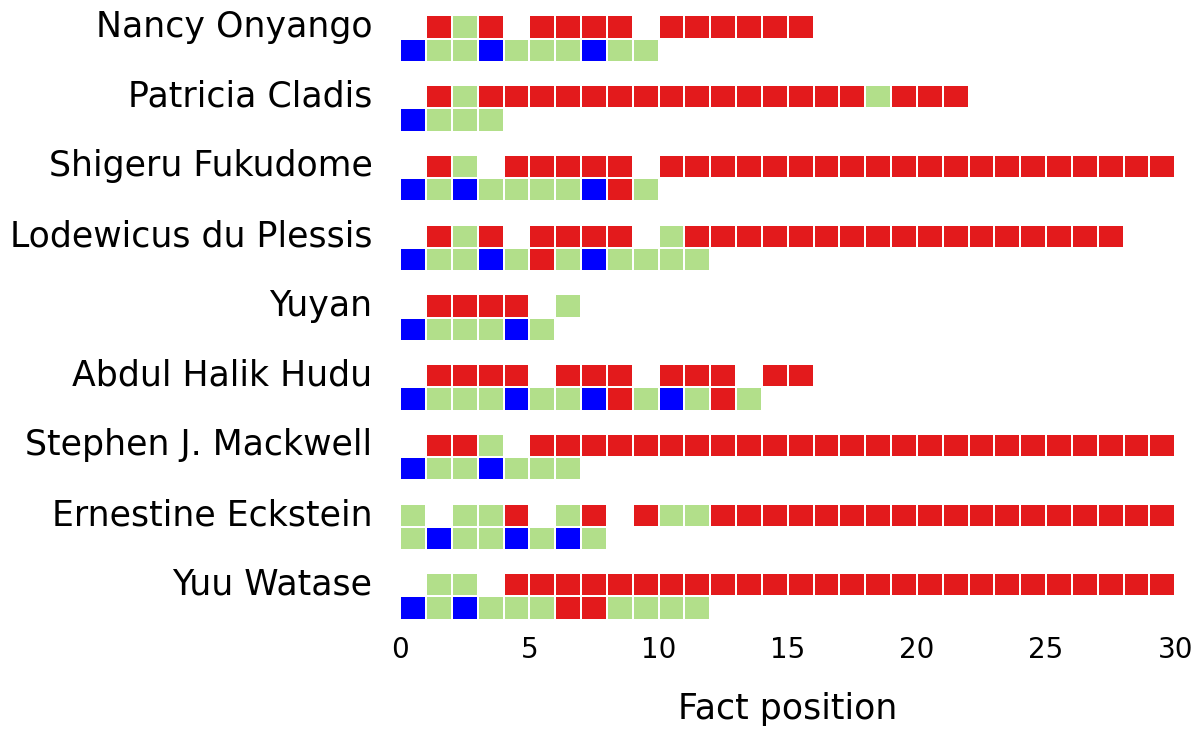}
    \caption{Examples of biographies that were most improved by adding QA calls. Each row represents a biography with two generated versions (one without QA calls and one with). Green~-- correct facts, red~-- incorrect facts, blue~-- API calls.}
    \label{fig:llama2-most-improved}
    \vspace{-2ex}
\end{figure}

\begin{table*}[ht]
\centering
  \small
  \begin{tabular}{lc|c|c|c|c|c}
    \toprule
         \bf Method & \bf Stop at & \bf toks/gen &\bf Fact &  \bf Triple &  \bf QAG&  \bf ROUGE-L\\
           &  &  & \bf Score&   \bf Score&  \bf  Score&  \\
         \toprule
         \bf Baseline & max tokens & 223.34 &19.45 & 10.57 & 30.91 & 4.17\\
         \hline
         \multicolumn{7}{l}{\bf Early stopping}\\
          & \texttt{EOS} in  top 5& \colorbox{oracle!40}{\phantom{0}60.33} &\colorbox{oracle!30}{12.72} & \colorbox{oracle!40}{\phantom{0}6.12} & \colorbox{internal!20}{36.60} & \colorbox{internal!40}{6.88}\\
          & SC-BERT incr. \textgreater0.5 & \colorbox{oracle!10}{184.84} &\colorbox{oracle!5}{19.05} & \colorbox{oracle!5}{10.35} & \colorbox{internal!10}{33.06}  &\colorbox{internal!20}{4.96}\\
          \hline
          \multicolumn{7}{l}{\bf Reranking (SC-BERT)}\\
          & max tokens & \colorbox{oracle!10}{189.60} & \colorbox{internal!5}{20.79} & \colorbox{internal!15}{11.71} & \colorbox{internal!20}{36.54} &\colorbox{internal!20}{5.04}\\
          & SC-BERT incr. \textgreater0.5& \colorbox{oracle!15}{157.67} & \colorbox{internal!5}{20.61} &\colorbox{internal!15}{11.77} &\colorbox{internal!20}{36.68} & \colorbox{internal!20}{5.11}\\\bottomrule
 \end{tabular}
  \caption{Factual accuracy for different generation strategies for Llama2-70B, when applied to the task in Section~\ref{sec:generalise}, of generating 5000 Wikipedia articles. Each score represents a measure of factual accuracy of the generated text with respect to the real Wikipedia article. All scores are calculated using the FactSumm pipeline \cite{factsumm}. }
  \label{tab:general}
  \vspace{-2ex}
\end{table*}

\paragraph{Results: the worst.} Surprisingly, using API calls gives close to no improvement: for similar number of facts per generation,  it is $10\%$ less accurate than other methods.
This is largely due to the model not handling errors of the API.  We found that adding more examples (and examples which ignore the API return) damaged the performance. The model makes many unnecessary calls, as it does not have an understanding of ``needing'' to retrieve information, rather it retrieves information whenever convenient. 
\par Adding calls in a few-shot manner poses a new challenge for semantic drift. We noticed that after generating a paragraph, the model would start a new paragraph about the API call (e.g., ``To make API calls use this method...'' or ``The API calls were executed at...''). In addition to being irrelevant, this is entirely hallucinated content. Removing this does not help significantly: it gives the SD score of $77.43\%$, which is only marginally lower than the baseline. For those samples which show most improvement in factuality, we note that the drift has been addressed by making many simple calls on almost every fact (Figure \ref{fig:llama2-most-improved}).

\section{Beyond Biographies} \label{sec:generalise}

\par The methods we presented can in principle  be applied to any type of text generation, not just biographical paragraphs. To showcase these capabilities, we apply them to writing any Wikipedia-style text. We prompt LLaMa2-70B to generate text about a topic (``This is a wikipedia article about \textit{topic}.'') and pass the generated text through the FactSumm pipeline \cite{factsumm}. 

\paragraph{Pipeline.}  The original goal of the pipeline is to measure factuality of a summary with respect to reference text. We re-purpose it to measure factuality of generated text with respect to the original Wikipedia article. We retrieve 5000 English Wikipedia articles\footnote{From \citet{wikidata}.} and calculate mean FactScore, TripleScore, QAG and ROUGE Score.

\paragraph{Evaluation.} FactScore and TripleScore extract trip\-lets (closed- and open-scheme, respectively) and score the overlap of these triplets between the reference and generated texts. For QAG Score, the module generates question-answer pairs based on the generated text, attempts to answer the questions based on the reference text and notes the number of identical answers. ROUGE calculates the similarity between the two texts based on n-grams matches. Together, all these scores paint a picture of the factuality of the generated text with respect to the Wikipedia article. We note that none of these metrics consider recall, but that we provide the average number of tokens generated per paragraph as a measure of information quantity.

\paragraph{Results: reranking helps again.} Table \ref{tab:general} shows that, in this more general setting, reranking yields higher factual accuracy at the cost of a reduction in generated facts. Therefore, this method has a positive impact on factual text generation beyond biographies. Here, we applied different metrics from FActScore* to assess the same phenomenon, offering a fresh perspective. Despite the different metrics, our methodology remained unaltered. The fact that the presented methods are robust to various metrics underscores their generality.

\section{Additional Related Work}

\par \textbf{Factual precision.} Recent surveys \cite{wang2023survey,rawte2023survey,Ji_2023} show that factuality evaluation has mostly been focused on short-form question answering, and improvements have largely been based on learning (pre-training, fine-tuning) or retrieval augmentation. Previous work \cite{Lee2022FactualityEL} attempts decoding-time enhancements, but reports these alone achieve factuality on-par with greedy decoding and concludes the need for training enhancements. Concurrent work \cite{chuang2023dola} contrasts various layers' logits. As opposed to SC-BERTScore methods, this requires access to model's internals and changes to inference code; futhermore it is not evaluated on long-form generation and restricted to one model class. Unlike other factuality enhancements, our methods do not directly fix incorrect facts, but use the semantic drift idea to inform when the model has ``ran out of correct facts''. They can be combined with any others to generate accurate and relevant text.

\paragraph{Semantic drift.} \citet{deng-etal-2022-model,cho-etal-2019-towards} characterize it linguistically via self-consistency, not truthfulness. Plausible and naturally flowing text would not be identified as drift.

\section{Conclusion}

By measuring the degree of separation between the correct and incorrect facts in the generated texts, we show that LLMs largely generate correct facts first and incorrect later. This lead us to methods that improve factual accuracy by stopping generation early. We show that even a simple method that encourages generating \texttt{EOS} leads to large improvements. This can further be improved by using a resample-then-rerank pipelines where for each sentence, we generate several versions and choose the best based on sentence similarity measures. Overall, our methods offer a practical compromise, balancing computation with performance, and build a foundation for further research. Importantly, they are directly applicable to any probabilistic auto-regressive language models.

\section{Limitations}
\paragraph{Model specifics.} We have applied our methods to LLaMa2-70B model and we trust that incentivising the \texttt{EOS} token and the SelfCheck-BERTScore methods  will work similarly well for other models. However, we note that the thresholds are likely not directly transferable to other models and that in order to employ similar strategies, model owners will have to tweak the thresholds to figure out the correct numbers for their case. 
\paragraph{Suitable tasks.} Even though our methods can be applied to any long-form text generation task, they are perhaps most relevant for tasks where factual accuracy is paramount (such as long form question answering or factual text generation). Early-stopping methods specifically are more suitable for tasks where generating false information is more harmful than not generating it (for example giving false medical advice). Our oracle for early-stopping removes 92\% of incorrect facts from the generated text, but this comes with the cost of removing 58\% of correct facts. These measurements (as can be seen in appendix \ref{app:F:metrics}) should be used for individual applications to debate trade-offs.
\paragraph{Textual diversity.} As this study is focused on factually-dense text, we did not take into account diversity of generated text, which may be relevant for more creative tasks such as story generation. For early stopping via sentence similarity, we chose to use SelfCheck-BERTScore which is sensitive to stylistic variations, as well as factual variations. However, there is no reason for which this metric cannot be replaced with other sentence similarity-metrics which account for style, thus retaining the creative factor of text generation. 
\paragraph{Automated evaluation.} We have used the FActScore pipeline, which is an automated evaluation pipeline for validating truthfulness of facts. We have validated the pipeline with human annotations (as detailed in Appendix \ref{sect_apdx:adapt_factscore_to_llama}), but as any automated pipeline it has an error margin. The reliability of the pipeline is heavily dependant on the reliability of its knowledge source, which in this case is Wikipedia -  one of the most commonly-used, accessible, large-scale, good quality, unstructured
knowledge sources \cite{Lee2022FactualityEL}.  
\paragraph{Future direction.} One can imagine many possible avenues of future directions for further understanding and mitigating semantic drift. For example, models could be further fine-tuned specifically to end generation when there is too much variability in the generation, critique models could be trained to identify the drift point based on model's internal states etc. We hope that with our work we have sufficiently highlighted the problem and set the first stepping stones for addressing it. 

\section{Acknowledgments}
\par We would like to thank Sewon Min for answering our questions about the FactScore \cite{min2023factscore} pipeline and accompanying code.

\nocite{*}

\bibliography{custom}

\appendix
\begin{appendices}

\section{Appendix A} \label{app:sd}

\subsection{Coherence, relevance and truthfulness}
The definition of semantic drift in Section \ref{sec:driftdef} states that the effects of semantic drift can be noted as a loss in coherence, relevance or truthfulness. The FActScore task described in Section \ref{sec:task} identifies all three categories of semantic drift effects. Here, we show examples of each. \\

\noindent \textbf{Coherence example}\\
Below an example of a generated biography, followed by the extracted facts together with their assigned labels. Inconsistency in the birth year is identified as False. 
\begin{algorithmic}
\small
    \State \texttt{\textbf{text:}}
    \State \texttt{Iggy Azalea (born 7 June 1990) is a rapper and singer. She was born in 1989.}
    \State \texttt{\textbf{facts:} }
    \State  \texttt{Iggy Azalea was born. (True)}
    \State \texttt{Iggy Azalea was born on June 7. (True)}
    \State  \texttt{Iggy Azalea was born on June 7,1990. (True)}
    \State  \texttt{Iggy Azalea. is a rapper. (True)}
    \State \texttt{Iggy Azalea is a singer. (True)}
    \State \texttt{She was born. (True)}
    \State \texttt{She was born in 1989. (False)}
\end{algorithmic}

\noindent Incoherence is probably one of the most studied types of semantic drift so far. Examples from literature include: 
\begin{algorithmic}
\small
    \State \texttt{ ``She had a large family and lived with her grandparents … In 1933 she gave birth to her first child … In July 1926, many of her friends attended her funeral'' \cite{liu2022tokenlevel}}
    \State \texttt{``Willie had too much stuff. Willie bought a shed to store all his stuff. Willie had a hard time putting up the shed. He called some friends for help. Willie sold his shed and and made enough money to pay for the house.'' \cite{wang2021sentence}}
\end{algorithmic} 

\noindent \\ 
\noindent \textbf{Relevance example}\\
As per our definition, a loss of relevance refers to the inclusion of irrelevant or redundant content. In the below example facts which are actually correct, but irrelevant to the context are labelled False. 
\begin{algorithmic}
\small
    \State \texttt{\textbf{text:}}
    \State \texttt{Iggy Azalea is a rapper and singer. Mariah Carey is a singer. Eminem is a singer. Bob Marley is also a singer.}
    \State \texttt{\textbf{facts:} }
    \State  \texttt{Iggy Azalea is a rapper. (True)}
    \State  \texttt{Iggy Azalea is a singer. (True)}
    \State  \texttt{Mariah Carey exists. (False)}
    \State  \texttt{Mariah Carey is a singer. (False)}
    \State \texttt{Eminem is a singer. (False)}
    \State  \texttt{Bob Marley is a singer. (False)}
\end{algorithmic}

\noindent \\ 

\noindent \textbf{Truthfulness example}
Truthfulness refers to the objective factuality of information, whether it is verifiable or not. In the example below, the scorer picks up on subtle inaccuracies (``Ignorant Art'' is a mixtape, not an album).

\begin{algorithmic}
\small
    \State \texttt{\textbf{text:}}
    \State \texttt{Iggy Azalea is a rapper from Melbourne, Australia. She is known for her hit single "Fancy" and her debut album Ignorant Art.}
    \State \texttt{\textbf{facts:} }
    \State  \texttt{Iggy Azalea is a rapper. (True)}
    \State \texttt{Iggy Azalea is from Melbourne, Australia. (True)}
    \State \texttt{Melbourne is a city in Australia (True).}
    \State \texttt{She is known for her hit single "Fancy". (True)}
    \State \texttt{She has a debut album called Ignorant Art. (False)}
    \State \texttt{"Fancy" is a hit single. (True)}
    \State \texttt{Ignorant Art is a debut album. (False)}
    \State \texttt{Ignorant Art is an album. (False)}
    
\end{algorithmic}

\subsection{Potential reasons for semantic drift}
Semantic drift is the term for a fairly broad phenomenon and there may be many reasons why it occurs. Some initial thoughts based on observations are: 
\begin{itemize}
    \item  Ambiguity: AI models may interpret ambiguous terms or phrases in ways that lead to a shift in the text's meaning.
    \item Loss of context: As text becomes longer, the model may lose track of the context.
    \item Digression: The AI model might include lengthy tangents or irrelevant information that detracts from the primary topic.
\end{itemize}

\subsection{SD Score and Purity}
\par The SD Score was inspired by purity measures in classification decision trees and can be seen as an edge-case calculation of purity. Recall that classification decision trees are made-up of nodes, where each node splits the training dataset into partitions using a criterion function on features of the data points. For each partition corresponding to a leaf node, the predicted class is the most common class in that partition. To find the right decision tree, we measure the purity of the partitions it creates. 
\par In the case of our task, the elements in the dataset are facts which have one feature: \textit{index} and are assigned a class: \textit{is\_supported}. The ``decision tree'' only has one split node (\textit{\textless index}) resulting in two leaf nodes (left-side and right-side). We then assign class 1 to every data point in the left-side and class 0 to every data point in the right-side. It is important to note that we always assign classes in this manner, regardless of which class is most common in the partition.

\section{Appendix B} \label{app:A}

\subsection{LLaMa2 70B Generation details}
\noindent We generate a maximum of 500 tokens with LLaMa2 70B model, with $temperature=0.6$ and $topp=0.9$. After generation we delete any unfinished sentences, as we found that the FActScore atomic fact extractor would hallucinate new facts when dealing with unfinished sentences. For analysis we also remove repetition, i.e. if the last sentence is repeated, then we remove it and stop generating. Our generated paragraphs have an average length of 255 tokens.

\subsection{FActScore Pipeline for LLaMa2}
\label{sect_apdx:adapt_factscore_to_llama}

We adapt the FActScore pipeline to rely on LLaMa2. The FActScore pipeline uses InstructGPT to extract atomic facts from input paragraphs. To do this, the model is given few-shot examples of atomic fact extraction. We use the same examples for LLaMa2 70B chat.  
To validate the performance of LLaMa2, we compare it against human annotations provided with the FActScore paper. The annotations consist of 180 paragraphs with extracted facts. We obtain a Pearson correlation coefficient of 0.94 between scores obtained from facts extracted by humans and scores obtained from facts extracted by LLaMa2.

\subsection{Impact of sampling strategy}
\begin{figure}[h]
    \centering
    \includegraphics[width=\linewidth]{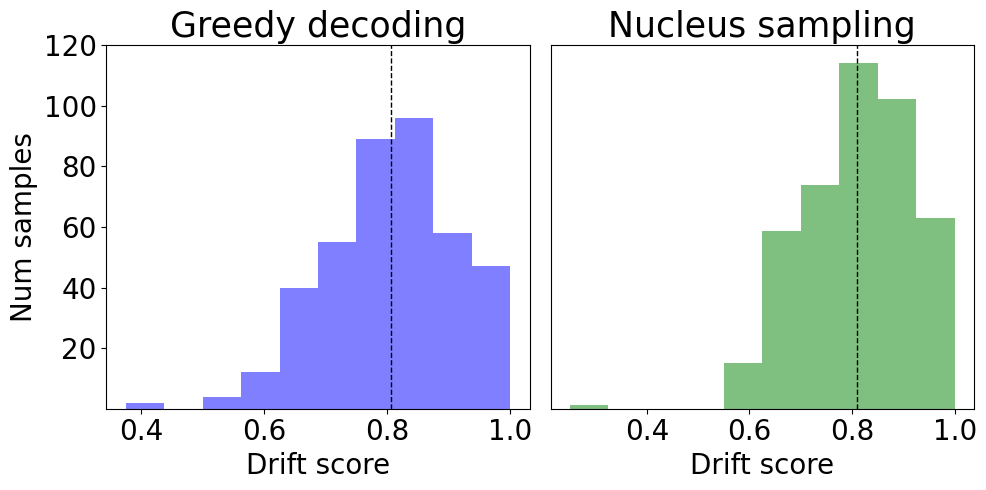}
    \caption{Comparison of SD score distribution in LLaMa2 70B based on decoding strategy.}
    \label{fig:decoding_startegy}
\end{figure}

As can be seen in Figure \ref{fig:decoding_startegy}, there is no considerable difference between the SD score obtained with greedy as opposed to nucleus sampling. We do note however, that the greedy-generated paragraphs tend to be more repetitive. 

\subsection{Truncation} \label{app:a:prune}
\begin{figure*}[t]
    \centering
    \includegraphics[width=\textwidth]{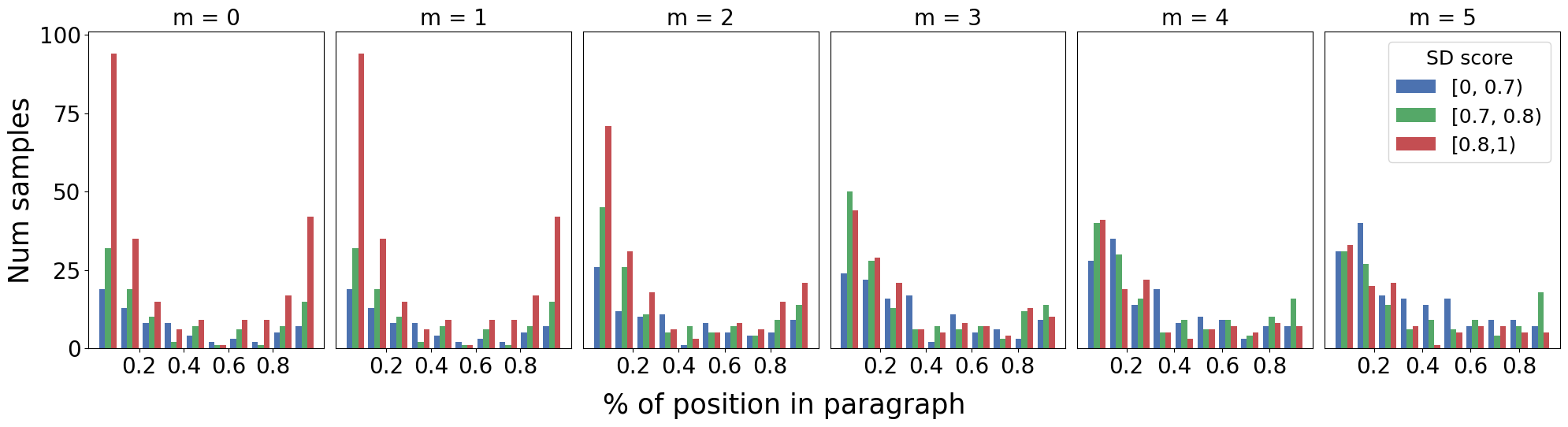}
    \caption{Drift position distribution after applying truncation varying the minimum number $m$ of facts on either side of the potential drift point, as described in Section \ref{sec:driftdef}.}
    \label{fig:pruning}
\end{figure*}

\noindent We define truncation in our particular case as described in Section~\ref{sec:driftdef}. We apply it in order to distinguish cases where the semantic drift high score is only caused by few samples to either side of the drift point. We experiment with $m \in [0,5]$ to see how the distribution of SD score and drift position are impacted. Results are in Figure \ref{fig:pruning}. We find that with $m=3$, there are 44.89\% paragraphs with SD score \textgreater 0.75.

\subsection{Statistical significance test} \label{app:a:stat}

For each paragraph with identified semantic drift (40\% of samples with SD score \textgreater = 0.75), we estimate the probability that the assigned SD score is due to chance. We shuffle the fact labels from the paragraph 1000 times and calculate the SD score for each shuffle. We find that, on average, higher or equal SD score is obtained in less than 0.02\% of shuffles\footnote{https://en.wikipedia.org/wiki/Permutation\_test}.

\subsection{Semantic drift identified examples}
\begin{figure}[h]
    \centering
    \includegraphics[width=\linewidth]{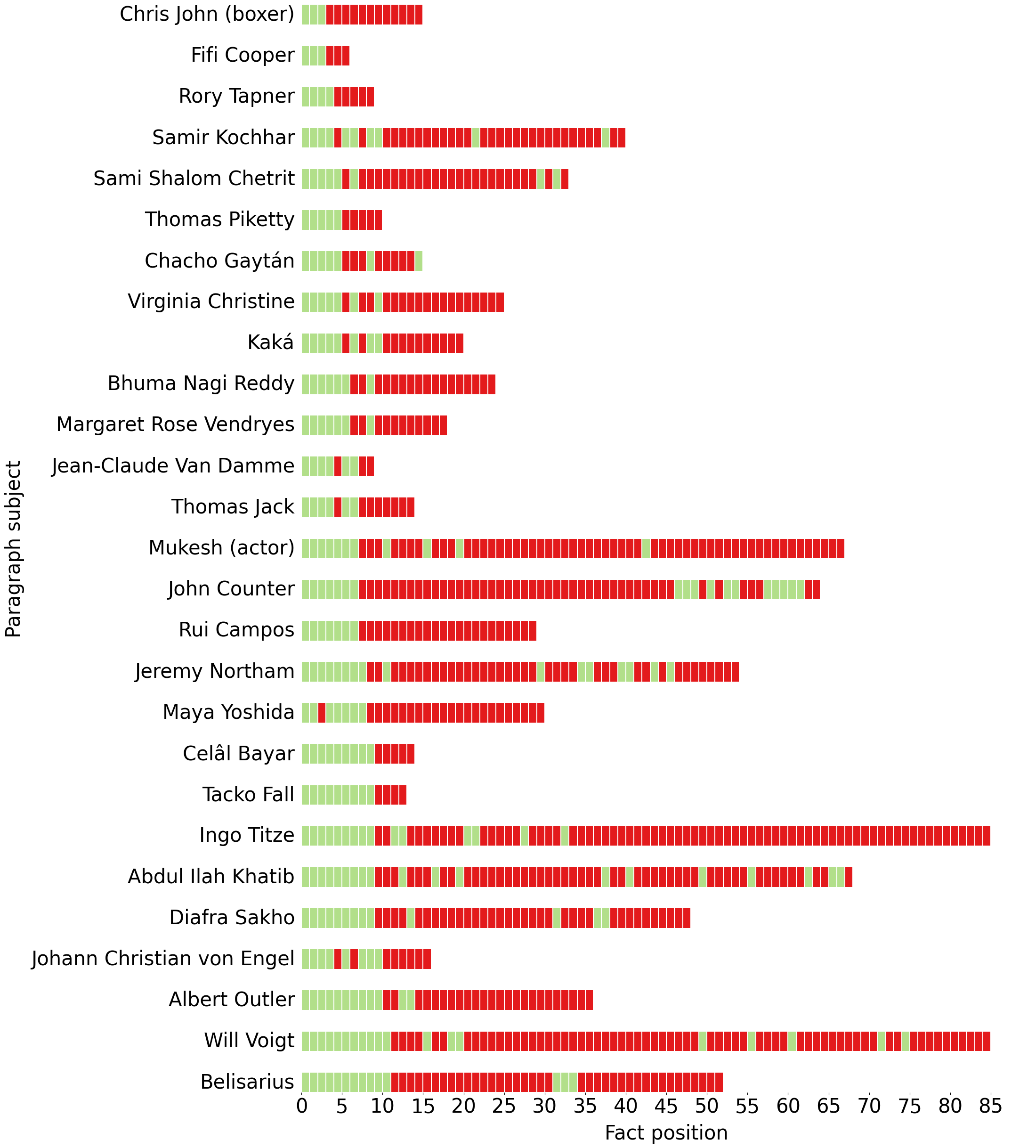}
    \caption{Examples of drifting paragraphs generated with LLaMa2 70B according to their SD score. Each row represents a paragraph where correct facts are in green, wrong facts are in red.}
    \label{fig:most_drifty}
\end{figure}

Figure \ref{fig:most_drifty} visually shows examples of paragraphs which display clear cases of semantic drift for various lengths of the paragraph.

\section{Appendix C} \label{app:B}

\subsection{Uncertainty metrics correlations}
\begin{figure}[h]
    \centering
    \includegraphics[width=\linewidth]{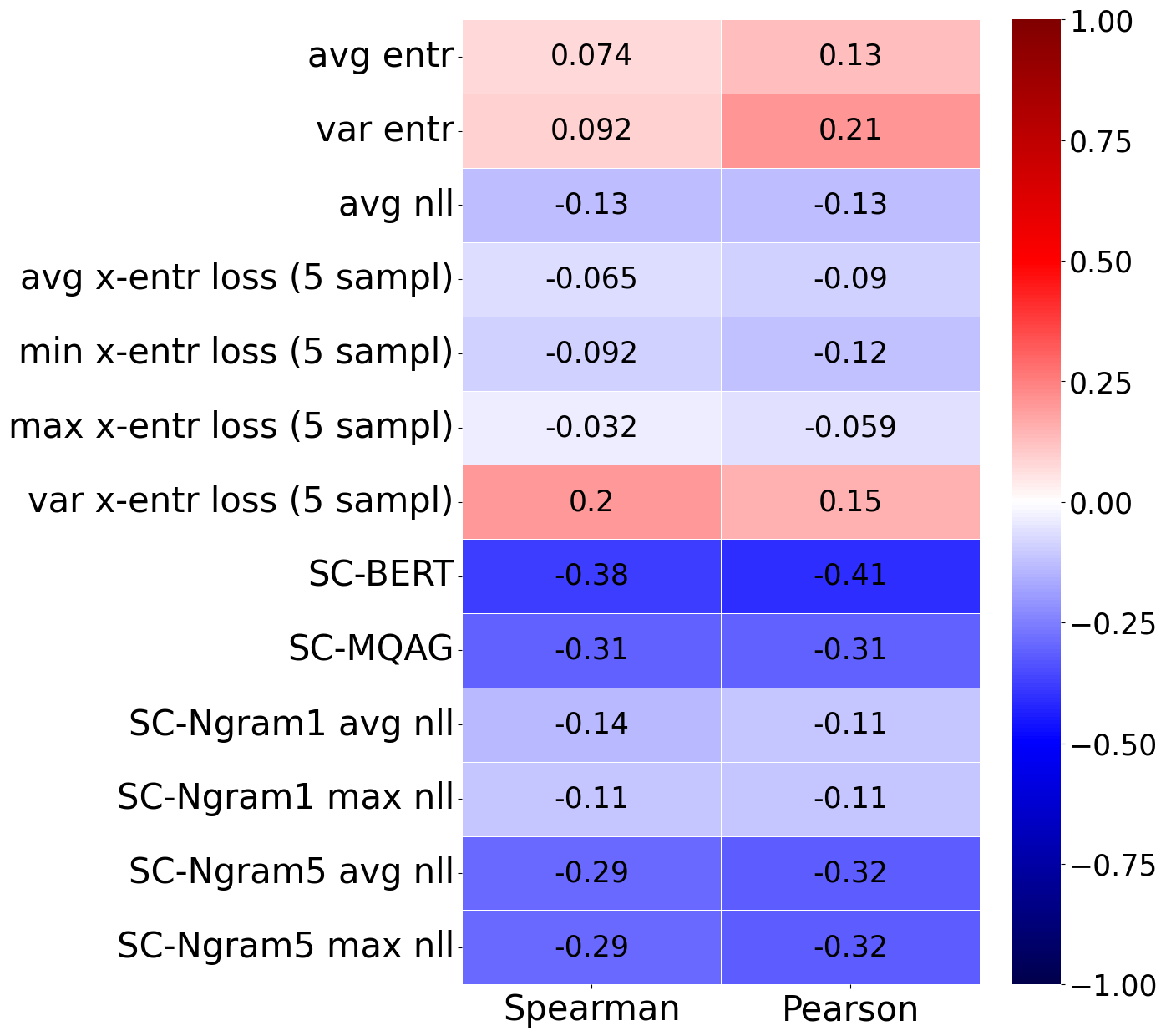}
    \caption{Correlation coefficients between all uncertainty metrics and fact accuracy. Calculated over 4516 sentences generated by LLaMa2 70B.}
    \label{fig:all_uncert}
\end{figure}

Figure \ref{fig:all_uncert} shows the correlation coefficients of all uncertainty metrics ran for our experiments. As mentioned in the main paper, the most significant correlation was for SelfCheck-BERTScore, calculated over 3 samples. 

\section{Appendix D} \label{app:C}
\subsection{SelfCheck-BERTScore}
For methods described in Section \ref{sect:early_stopping}, we conducted more experiments to determine how threshold $T$ on the relative increase in SelfCheck-BERTScore should be chosen and whether it could be an absolute value threshold, as opposed to a relative increase value. We also provide more details for how the paragraph samples were generated. 

\subsection{Absolute thresholds for ending generation}
\noindent A more naive method for stopping generation using SelfCheck-BERTScore is to simply threshold the absolute value of the score and stop generation whenever the score crosses the threshold. When applying the method, we found that many biographies would actually begin with a first sentence above the threshold, thus resulting in empty paragraphs for any value of the threshold sufficiently low to be useful. We show the results of either keeping or deleting those biographies which begin above the threshold in Table \ref{tab:eosbert_absoluteval}.

\par
\begin{figure}[hb]
    \centering
    \includegraphics[width=\linewidth]{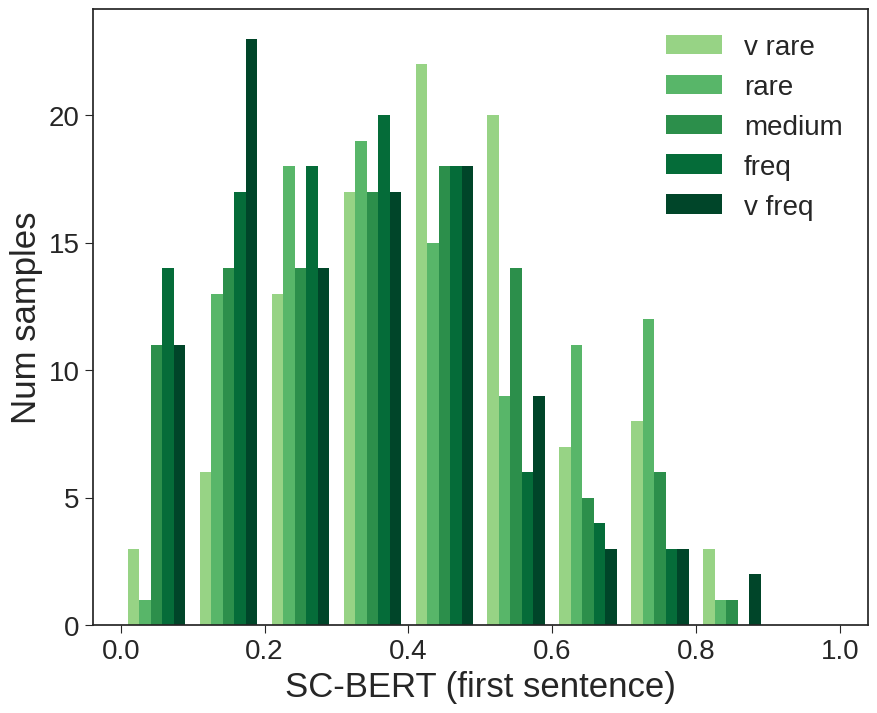}
    \caption{Distribution of SelfCheck-BERTScore for first sentence in paragraph, by popularity of topic.}
    \label{fig:bertsc_distrib}
\end{figure}

 An interesting corollary finding is the distribution of SelfCheck-BERTScore in the first sentence by popularity class of the topic. The paragraphs which have highest SelfCheck-BERTScore in the first sentence are those with lower popularity, and consequently those for which the above method would not generate any facts (Figure \ref{fig:bertsc_distrib}). 

\subsection{Number of sampled paragraphs}
\noindent The calculation of the SelfCheck-BERTScore hinges on using $N$ sampled paragraphs. Each sentence in the original paragraph is scored based on its BERTScore with respect to each sampled paragraph. We experimented with $N \in \{ 1, 3, 5, 10, 100\}$. We found only marginal improvements for $N>5$ and that the improvements are more visible when using smaller $T$. We also experimented with generating $N$ paragraph samples with a temperature setting of 1. As the original paper \citet{Manakul2023SelfCheckGPTZB} suggests, high temperature  should result in more accurate SelfCheck-BERTScore. However, we find the improvements on $temperature=1$ to be marginal from $temperature=0.6$, $topp=0.9$.

\subsection{Rerank past-early stopping point}
\par One interesting experiment, but which did not yield satisfactory results, was to use the early-stopping strategy described in Section \ref{sect:early_stopping} and combine it with the reranking strategy from Section \ref{sec:rerank} by resampling-and-reranking only past the early-stopping point. The hope was that we can extend current paragraphs by adding more correct facts. We found that we could extend the paragraphs with an average of 2.12 facts per generated paragraph, but that this came with a loss of factuality of 1.67\%.

\section{Appendix E} \label{app:D}
\subsection{Inference with API call} \label{app:D:example}
Below is an example illustrating the inference flow for generating text with embedded API calls. It consists mainly of two prompts: one for defining how to make the API call and one for integrating the response of the API call. When finished with inference step 2, we remove the API call from the generated text and repeat the same flow again with the previously generated text as the new prompt we want to complete. \\

\begin{algorithmic}
\small
    \State \textbf{Prompt 1:}
    \State \texttt{Your task is to add calls to a Question Answering API to a piece of text. The questions should help you get information required to complete the text. You can call the API by writing [QA(question)] where question is the question you want to ask. Here are some examples of API calls:}
    \State \texttt{Joe Biden was born in [QA(Where was Joe Biden born?)]}
    \State \texttt{This is a Wikipedia article about Napoleon. Napoleon }\\
    \State \textbf{Inference 1:}
    \State \texttt{was born in [QA(Where was Napoleon born?)]}\\ 
    \State \textbf{Execute API call.}\\
    \State \textbf{Prompt 2:}
    \State \texttt{Your task is to complete a piece of text, by using answers from an API call. APIs are called by writing [QA(question) -> answer] where question is what was sent to the API and answer is the response. Here are some examples of texts with API calls:}
    \State \texttt{Joe Biden was born in [QA(Where was Joe Biden born?) -> Scranton] Scranton, Pennsylvania.}
    \State \texttt{Napoleon was born in [QA(Where was Napoleon born?) -> Ajaccio]}\\ 

    \State \textbf{Inference 2:}
    \State \texttt{was born in [QA(Where was Napoleon born?) -> Ajaccio] Ajaccio, Corsica.}\\ 

    \State \textbf{Repeat.} 
\end{algorithmic} 

\section{Appendix F} \label{app:F}
\subsection{Factual precision and recall metrics} \label{app:F:metrics}
Because FActScore is a precision-focused metric, to get a better idea of the impact of each regeneration strategy and each early-stopping strategy, we provide more metrics in Table \ref{tab:prec_rec}. The recall on incorrect facts shows the percentage of incorrect facts that were present in the original generation, then removed by the early stopping method. 

\begin{table}[h]
\begin{tabular}{l|ll|ll}
\multirow{2}{*}{Method}                                                               & \multicolumn{2}{l|}{Incorrect facts}                 & \multicolumn{2}{l}{Correct facts}                    \\ \cline{2-5} 
                                                                                      & \multicolumn{1}{l|}{Prec.}      & Rec.         & \multicolumn{1}{l|}{Prec.}      & Rec.         \\ \hline
baseline                                                                              & \multicolumn{1}{l|}{}               &                & \multicolumn{1}{l|}{44.56}          &                \\ \hline
oracle                                                                                & \multicolumn{1}{l|}{66.39}          & \textbf{92.47} & \multicolumn{1}{l|}{\textbf{81.68}} & 41.76          \\
eos\_top\_5                                                                           & \multicolumn{1}{l|}{61.99}          & 75.1           & \multicolumn{1}{l|}{57.96}          & 42.71          \\
eos\_top\_10                                                                          & \multicolumn{1}{l|}{58.94}          & 93.58          & \multicolumn{1}{l|}{70.29}          & 18.90          \\
SC-Bert \textgreater{}.7                                                              & \multicolumn{1}{l|}{\textbf{69.03}} & 74.44          & \multicolumn{1}{l|}{64.76}          & 58.44          \\
SC-Bert \textgreater{}.5                                                              & \multicolumn{1}{l|}{65.57}          & 82.19          & \multicolumn{1}{l|}{67.63}          & 46.30          \\
SC-Bert \textgreater{}.3                                                              & \multicolumn{1}{l|}{62.69}          & 88.19          & \multicolumn{1}{l|}{70.24}          & 34.69          \\ \hline
re-rank                                                                               & \multicolumn{1}{l|}{}               &                & \multicolumn{1}{l|}{53.27}          &                \\
\begin{tabular}[c]{@{}l@{}}rerank + \\ SC-Bert \textgreater{}.7\end{tabular}          & \multicolumn{1}{l|}{60.35}          & 56.07          & \multicolumn{1}{l|}{63.72}          & \textbf{67.67} \\
\begin{tabular}[c]{@{}l@{}}rerank + \\ SC-Bert \textgreater{}.5\end{tabular}          & \multicolumn{1}{l|}{57.04}          & 70.1           & \multicolumn{1}{l|}{67.18}          & 53.69          \\
\begin{tabular}[c]{@{}l@{}}rerank + \\ SC-Bert \textgreater{}.3\end{tabular} & \multicolumn{1}{l|}{54}             & 82.1           & \multicolumn{1}{l|}{71.11}          & 38.64          \\ \hline
1 API call                                                                            & \multicolumn{1}{l|}{}               &                & \multicolumn{1}{l|}{43.93}          &                \\
inf API calls                                                                         & \multicolumn{1}{l|}{}               &                & \multicolumn{1}{l|}{54.42}          &               
\end{tabular}
\caption{Metrics for incorrect facts, showing precision (\%incorrect facts out of those removed by early stopping), recall (\%incorrect facts removed); and for correct facts, showing precision (FActScore) and recall (\%remaining correct facts). For each of the generation strategies, the metrics are calculated with respect to the base generation.}
\label{tab:prec_rec}
\end{table}

\begin{table*}[ht]
  \small
  \begin{tabularx}{\textwidth}{|p{4cm}|p{1.4cm}|p{1.8cm}|p{1.8cm}|X|X|X|X|}
    \hline
    \textbf{Early stopping (threshold $T$)} & \textbf{if $S_{0} > T$} & \textbf{facts /gen} & \textbf{No answer} & \textbf{FAct Score* (\%)} & \textbf{SD score (\%)}  \\
    \hline
    SC-BERTScore  \textgreater 0.8 & keep $S_{0}$& 28 &1 & 52.73 & 75.77 \\
    SC-BERTScore \textgreater 0.5 & keep $S_{0}$& 16.21 & 1& 66.05 & 66.87 \\
    SC-BERTScore  \textgreater 0.2 & keep $S_{0}$& 21.92 & 1& 62.87 & 67.63 \\
    \hline
    SC-BERTScore \textgreater 0.8 & delete $S_{0}$ &21.26 &17 & 58.07 & 74.18 \\
    SC-BERTScore  \textgreater 0.5 & delete $S_{0}$&4.64 & 272 & 88.66 & 66.87 \\
    SC-BERTScore  \textgreater 0.2 & delete $S_{0}$&4 & 493 & 92.85 & 58.64 \\
    \hline
    @drift point & n/a &10 & 1 & 81.68& 47.94 \\
    \hline
  \end{tabularx}
  \caption{Comparing FActScore* and SD score for LLaMa2 70B cutting generation based on SelfCheckBERT Score threshold $T$. The second column shows behaviour in the case in which the first sentence of the generation already exceeds threshold $T$. ``No answer'' shows the number of paragraphs (out of the total 500) for which the model produces no facts. }
  \label{tab:eosbert_absoluteval}
\end{table*}

\end{appendices}

\end{document}